\documentclass[sigconf]{acmart}

\usepackage{multirow} % multi row cells
\usepackage{array}

\AtBeginDocument{%
  \providecommand\BibTeX{{%
    \normalfont B\kern-0.5em{\scshape i\kern-0.25em b}\kern-0.8em\TeX}}}

\acmConference[MAD '24]{M3rd ACM International Workshop on Multimedia AI against Disinformation organized with the ACM International Conference on Multimedia Retrieval (ICMR '24)}{June 10--13, 2024}{Phuket, Thailand}

\copyrightyear{2024}
\acmYear{2024}
\setcopyright{acmlicensed}\acmConference[MAD '24]{3rd ACM International
Workshop on Multimedia AI against Disinformation}{June 10, 2024}{Phuket,
Thailand}
\acmBooktitle{3rd ACM International Workshop on Multimedia AI against
Disinformation (MAD '24), June 10, 2024, Phuket, Thailand}
\acmDOI{10.1145/3643491.3660287}
\acmISBN{979-8-4007-0552-6/24/06}

\newcolumntype{L}[1]{>{\raggedright\let\newline\\\arraybackslash\hspace{0pt}}m{#1}}
\newcolumntype{D}[1]{>{\centering\let\newline\\\arraybackslash\hspace{0pt}}m{#1}}

\newcommand{\vaddfull}{Visual-Audio Discrepancy Detection}
\newcommand{\vadd}{VADD}
\newcommand{\taud}{TAU}
\newcommand{\pristine}{unmodified}
\newcommand{\Pristine}{Unmodified}

\begin{document}

\title{Visual and audio scene classification for detecting discrepancies in video: a baseline method and experimental protocol}

\author{Konstantinos Apostolidis}
\email{kapost@iti.gr}
\orcid{0000-0002-9470-6332}
\affiliation{
    \institution{Information Technologies Institute, CERTH}
    \city{Thessaloniki}
    \country{Greece}
}

\author{Jakob Abe{\ss}er}
\email{jakob.abesser@idmt.fraunhofer.de}
\orcid{0000-0003-4689-7944}
\affiliation{
    \institution{Fraunhofer Institute for Digital Media Technology}
    \city{Ilmenau}
    \country{Germany}
}

\author{Luca Cuccovillo}
\email{luca.cuccovillo@idmt.fraunhofer.de}
\orcid{0000-0001-5559-6508}
\affiliation{
    \institution{Fraunhofer Institute for Digital Media Technology}
    \city{Ilmenau}
    \country{Germany}
}

\author{Vasileios Mezaris}
\email{bmezaris@iti.gr}
\orcid{0000-0002-0121-4364}
\affiliation{
    \institution{Information Technologies Institute, CERTH}
    \city{Thessaloniki}
    \country{Greece}
}

\renewcommand{\shortauthors}{  }
\renewcommand{\shorttitle}{  }

\begin{abstract}
This paper presents a baseline approach and an experimental protocol for a specific content verification problem: detecting discrepancies between the audio and video modalities in multimedia content. We first design and optimize an audio-visual scene classifier, to compare with existing classification baselines that use both modalities. Then, by applying this classifier separately to the audio and the visual modality, we can detect scene-class inconsistencies between them. To facilitate further research and provide a common evaluation platform, we introduce an experimental protocol and a benchmark dataset simulating such inconsistencies. Our approach achieves state-of-the-art results in scene classification and promising outcomes in audio-visual discrepancies detection, highlighting its potential in content verification applications.
\end{abstract}




% \received{15 March 2024}
% \received[revised]{20 March 2024}
% \received[accepted]{5 June 2024}

\maketitle

\section{Introduction}
Digital disinformation, the intentional dissemination of false or misleading information through digital media, encompasses various deceptive tactics, including fabricated news, tampered images, manipulated videos, and misleading narratives. Professionals across fields, such as journalists, security experts, and emergency management officers, are increasingly concerned about discerning genuine content from manipulated material. In response, AI-based content verification tools and integrated toolboxes like~\cite{teyssou2017invid} have emerged to assess the authenticity and integrity of digital content. While manual verification is labor-intensive, such support tools driven by machine learning offer scalable solutions, allowing rapid multimedia analysis and aiding in prioritizing human effort. 

In audio editing, incorporating the soundscape of an acoustic scene has become standard practice. This soundscape addition serves two main purposes: first, it helps to mask editing points, seamlessly blending different audio segments; second, it enhances immersion, providing viewers with a more engaging auditory experience. For instance, in the context of news broadcasts or documentary filmmaking, audio soundscapes are often captured at the event being depicted in the video. 

However, when malicious users attempt to manipulate multimedia content, they are unlikely to have access to authentic audio soundscapes from the actual events. Instead, they may resort to using pre-existing ambient sounds, which may not align with the visual content they are manipulating. As a result, the manipulated content produced by malicious users could contain inconsistencies between the audio and video modalities. These inconsistencies could range from subtle discrepancies in background noise to more obvious mismatches in environmental cues, leading to a loss of the video's credibility.

While detecting AI-generated fabrications has garnered attention, identifying subtle but crucial disparities in audio-visual streams remains unexplored. Content verification often overlooks inconsistencies between different modalities, such as between the audio and visual components of video. Such disparities, whether it is an audio track incongruous with the visual scene, or the presence of conflicting environmental cues, may signal that an incoming video has been fabricated. While certain instances of audio-visual mismatches identified by our algorithm may be easily perceptible to human observers, our proposed method proves particularly valuable in automatically processing extensive raw video corpora, providing discerning insights into potential tampering occurrences. 

\begin{figure}[ht]
    \centering
    \includegraphics[trim={0 0cm 0cm 0},clip,width=0.99\linewidth]{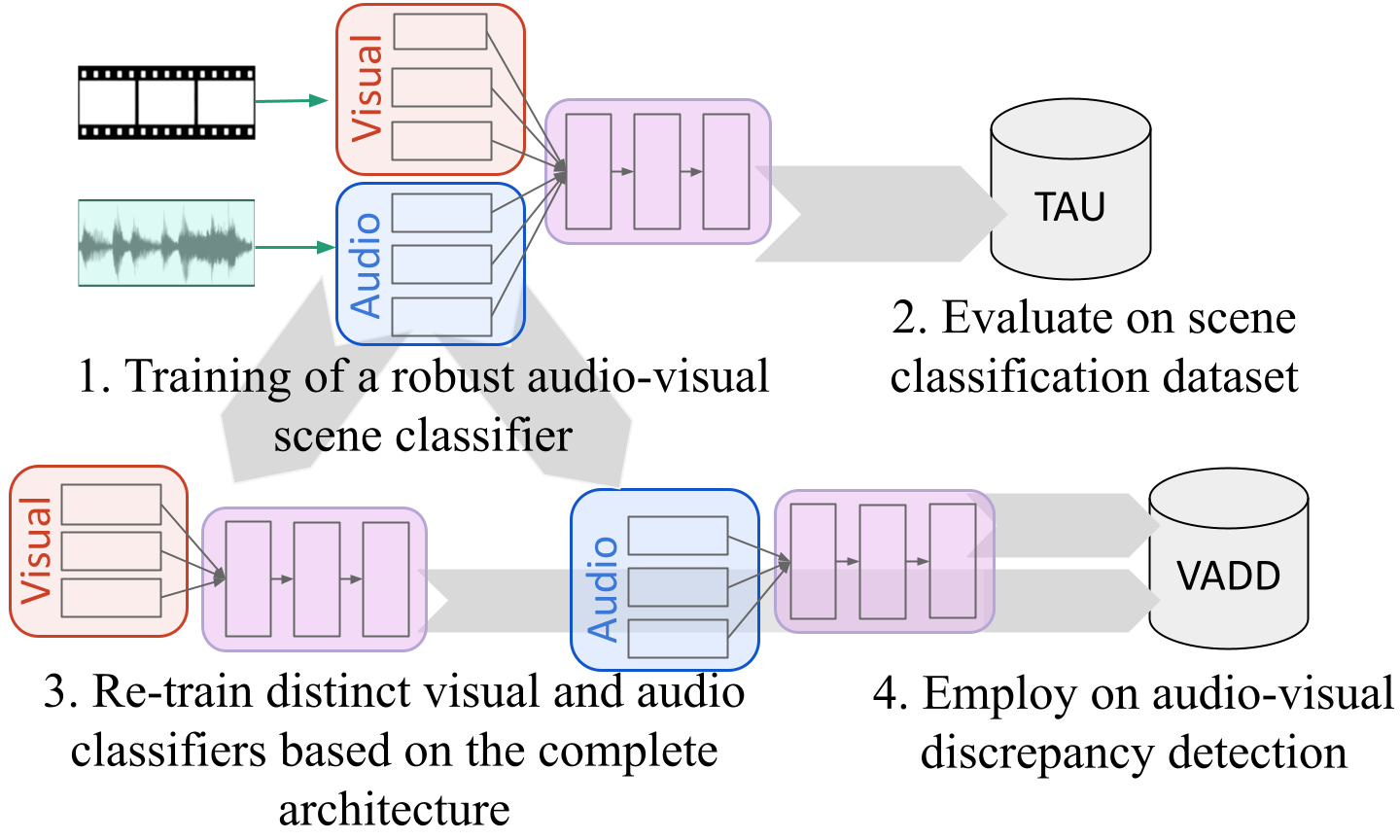}
    \caption{The overall procedure employed in this paper. The red blocks represent the ensemble of visual embeddings (three blank rectangles inside). The blue blocks represent the ensemble of audio embeddings (three blank rectangles inside).} 
    \label{fig:overall}
\end{figure}

Our proposed baseline method adapts visual- and audio-scene classification techniques to detect such discrepancies, as depicted in Fig.~\ref{fig:overall}. Another key contribution of this paper is the introduction of an experimental protocol and a benchmark dataset tailored to this task. We freely provide this dataset, as well as the source code of the proposed baseline method to serve as a valuable resource for the scientific community, promoting further advancements in the field of content verification.

The rest of this paper is organized as follows: Section~\ref{sec:soa} provides an overview of relevant work in audio and visual scene classification and methods related to visual-audio discrepancies detection. Section~\ref{sec:data} presents the process of generating our benchmark dataset and its key attributes. Section~\ref{sec:proposed} describes our methodology for detecting discrepancies using audio and visual scene classifiers and the proposed aggregation scheme. Section~\ref{sec:exps} details our experiments on scene classification, visual-audio discrepancies detection, and a short ablation study for our proposed classifier, followed by our conclusions in Section~\ref{sec:conclusions}.

\section{Related Work}
\label{sec:soa}

\subsection{Acoustic Scene Classification}
\label{sec:asc}
Acoustic Scene Classification (ASC) is an audio tagging task in which an audio recording is classified into a semantic scene category based on the location of the recording. While acoustic scenes are often characterized by some unique sound events (such as car and truck sounds in the acoustic scene ``street traffic''), some groups of scene classes share many types of sounds (such as the scene classes ``pedestrian street'' and ``public square''), making their classification often challenging and ambiguous. The annual Detection and Classification of Acoustic Scenes and Events (DCASE) challenges and workshops have strongly stimulated research in the field of ASC over the last 10 years.

Early ASC methods combined traditional feature representations such as Mel-Frequency Cepstral Coefficients (MFCC) and classification approaches such as Gaussian Mixture Models (GMM) or Support Vector Machines (SVM). The current state-of-the-art models are instead based on deep neural network architectures such as convolutional neural networks or ResNets, and process audio data represented either as raw audio samples or as time-frequency representations such as Mel-spectrograms \cite{Virtanen_2018_SoundScenesAndEvents_BOOK}. In contrast to the sound event detection (SED) task, where sounds need to be localized in time, ASC models usually have a final pooling layer to obtain a time-aggregated classification result. Two of the most current research trends in the field of ASC are the development of domain adaptation methods to cope with microphone mismatch scenarios \cite{Drossos_2019_UnsupervisedDomainAdaptation_WASPAA,Mezza_2021_BWSM_EUSIPCO,Johnson_2021_ISANormalization_EUSIPCO} and the design of resource-efficient deep neural network architectures, which enable the deployment of ASC algorithms on mobile devices and hearables \cite{Kim_2021_ResidualNormalizationASC_DCASE}. An extensive literature survey on ASC was published in \cite{Abesser_2020_ascOverview_mdpi}.

\subsection{Visual Scene Classification}
\label{sec:vsc}
Visual Scene Classification (VSC) categorizes images or videos into predefined scene classes based on visual content, often relying on image classification techniques. Deep learning has revolutionized image classification by automatically learning complex features from data. Influential architectures like Residual Network (ResNet)~\cite{he2016deep}, DenseNet~\cite{huang2017densely}, and EfficientNet~\cite{tan2019efficientnet} optimize model depth, width, and resolution. More recently, vision transformers \cite{dosovitskiy2020image} have built upon the idea of adapting the transformer architecture, originally designed for natural language processing tasks, to the field of image classification. The image is broken into fixed-size patches and treated as tokens, enabling the application of self-attention mechanisms. It has gained significant attention for its competitive performance on various benchmark datasets. Finally, contrastive language-image models aim to learn joint representations of textual and visual data. These models align image and text representations in a shared latent space, bringing similar image-text pairs closer while pushing dissimilar ones apart. One notable example is CLIP (Contrastive Language-Image Pre-training) \cite{radford2021learning}, which has been successfully used for zero-shot and few-shot image classification, without explicit task-specific training as in \cite{wang2023exploring}.

Most of the discussed architectures have readily available pre-trained models on large image classification datasets like ImageNet and Places365. Places365 is particularly noteworthy, as it is extensively utilized in scene classification research. This dataset provides a rich and diverse assortment of images depicting various indoor and outdoor scenes from across the world. Each image is labeled with a specific scene category, such as natural landscapes, urban environments, interiors, and outdoor spaces. In total, the dataset includes 365 scene categories and comprises 1.8 million images.

\subsection{Visual-audio discrepancies detection}
To our knowledge, the sole existing work addressing similar concerns is the \cite{bolles2017spotting} study, which dedicates a section to detecting scene discrepancies between audio and visual streams. However, their approach involves the independent design of audio and visual scene detectors, lacking evaluation on a standardized benchmark dataset. Furthermore, the evaluation relies on synthetic media constructed from video items of a publicly available dataset, forming a very limited testing set size (100 videos) and an undisclosed evaluation protocol. Additionally, the study focuses on a binary classification of audio-visual scenes, categorizing them as either ``outdoors'' or ``indoors''. A closely related work to this domain is Task 1B of the DCASE 2021’s challenge which involves categorizing videos based on their audio and visual content. Participants in this challenge are required to develop and evaluate systems that can jointly analyze audio and visual information to determine the type of scene depicted in multimedia recordings. The performance of participating systems is evaluated based on classification accuracy on the provided TAU Urban Acoustic Scenes 2020 Mobile challenge~\cite{Wang2021_ICASSP} dataset - from now on we will refer to this dataset simply as {\taud}.

The top-performing method of DCASE 2021 Task 1B, as discussed in \cite{wang2021audio}, leverages transfer learning and a hybrid fusion strategy. Its authors employ pre-trained deep neural networks for extracting both audio and visual features from videos. These features are combined using a hybrid fusion strategy, encompassing both early fusion (combining features before classification) and late fusion (combining classification results from separate audio and visual models).

\section{Dataset and Experimental Protocol}
\label{sec:data}

\subsection{Original scene classification dataset}
The {\taud} dataset focuses on Audio-Visual Scene Classification (AVSC), featuring audio and video recordings from various urban locations. It includes 10 urban acoustic scene classes: ``bus'', ``metro'', ``street pedestrian'', ``public square'', ``street traffic'', ``tram'', ``park'', ``airport'', ``shopping mall'', and ``metro station''. Each class has audio and video recordings. Audio is in mono, 48 kHz, 24-bit WAV format, while video is synchronized MP4 clips at 25 fps, with resolutions ranging from 480p to 1080p.
The dataset is divided into \textit{Development} and \textit{Evaluation} sets. The \textit{Development} set consists of 12,291 one-shot videos (i.e., a single continuous video shot, without any cuts, edits, or interruptions), each 10 seconds long, divided into training and testing portions. The \textit{Evaluation} set contains 72,000 videos, each lasting 2 seconds. Annotations for the \textit{Evaluation} set are withheld and were managed by DCASE2021 organizers solely for challenge participants.

\subsection{Proposed visual-audio discrepancies experimental protocol}
We introduce the {\vaddfull} ({\vadd}) experimental protocol, curated to facilitate research in detecting discrepancies between visual and audio streams in videos. The dataset includes a subset of videos in which the visual content portrays one class (e.g., an outdoor scenery), while the accompanying audio track is sourced from a different class (e.g., the sound of an indoor environment). Aiming to leverage the wealth of visual and auditory data already available in the already existing {\taud} dataset, additionally expediting the data collection process, our {\vadd} experimental protocol and dataset is created by re-purposing data and providing annotations for the {\taud} dataset. Specifically, we swap the audio and video streams for half of the videos in the {\taud} dataset, to create ``manipulated'' samples while keeping the rest of the videos unchanged to have ``\pristine'' samples. We ensured balanced sets of \pristine and manipulated samples through the following process (see Table~\ref{tab:shuffling_10} for the resulting distribution of samples across classes):

\begin{enumerate}
	\item Randomly select half of each class's samples for inclusion in the ``\pristine'' set, and put the remaining half in a ``bucket'';
	\item Randomly select from the bucket two items belonging to different classes, mutually swap their audio streams so that two new audio-visual samples are generated, and add them to the ``manipulated'' set;
	\item Repeat step 2 until all items left in the bucket belong to the same class;
	\item Finally, add these remaining items to the ``\pristine'' set.
\end{enumerate}

\begin{table}
	\centering
	\caption{Class-wise Distribution of \Pristine and Manipulated Samples in the {\vadd} dataset}
	\label{tab:shuffling_10}
	\begin{tabular}{lccc}
    	\hline
    	Class           	& Total &      	\Pristine (\%) &  Manipulated  	(\%) \\
    	\hline
    	airport         	&   281 &      	141 (50.18\%) &      	140 (49.82\%) \\
    	bus             	&   327 &      	164 (50.15\%) &      	163 (49.85\%) \\
    	metro           	&   360 &      	180 (50.00\%) &      	180 (50.00\%) \\
    	metro station   	&   386 &      	193 (50.00\%) &      	193 (50.00\%) \\
    	park            	&   386 &      	193 (50.00\%) &      	193 (50.00\%) \\
    	public square   	&   387 &      	194 (50.13\%) &      	193 (49.87\%) \\
    	shopping mall   	&   387 &      	194 (50.13\%) &      	193 (49.87\%) \\
    	street pedestrian   &   421 &      	211 (50.12\%) &      	210 (49.88\%) \\
    	street traffic  	&   402 &      	201 (50.00\%) &      	201 (50.00\%) \\
    	tram            	&   308 &      	154 (50.00\%) &      	154 (50.00\%) \\
    	\hline
                       	&  3645 &     	1825 (50.07\%) &     	1820 (49.93\%) \\
    	\hline
	\end{tabular}
\end{table}

We introduce a 3-class version of our dataset, derived from the original 10-class dataset, motivated by the need to provide two levels of difficulty. This 3-class taxonomy aligns with the ``Low-Complexity Acoustic Scene Classification'' sub-task of the ``Acoustic scene classification'' challenge, a simplification of the acoustic scene classification task, where the 10 acoustic scene classes are mapped to three classes: indoor, outdoor, and vehicle. In this 3-class variant of {\vadd}, videos are categorized into higher-level scene classes, simplifying the task by condensing the number of potential discrepancies. Conversely, the 10-class variant offers a more intricate challenge, reflecting the diversity of discrepancies in more realistic scenarios. 

The goal of our dataset is to enable the evaluation of methods that can detect discrepancies between the visual and audio streams. To assess the effectiveness of such methods in identifying manipulated samples from the {\vadd} dataset, we propose using the F1-score of Precision and Recall. We provide this dataset in our GitHub repository\footnote{https://github.com/IDT-ITI/Visual-Audio-Discrepancy-Detection}, i.e. the list of {\taud}'s videos that belong to the \pristine samples set of the {\vadd} dataset, and the tuples of {\taud}'s videos for which the audio should be swapped to create the set of manipulated videos.

\section{Proposed Method}
\label{sec:proposed}

We have developed a methodology for detecting discrepancies between audio and visual elements in video content. Initially, we train a robust joint audio-visual classifier specifically designed for scene classification tasks, utilizing both audio and visual modalities. This training utilizes the training portion of the development set from the {\taud} dataset. Subsequently, the performance of the joint classifier is evaluated on the test portion of the development set from the {\taud} dataset to ensure its efficacy in scene classification tasks. Following the selection of the architecture and the evaluation of the scene classifier's effectiveness, we retrain separate classifiers for each modality. These classifiers are intended for use individually on the audio and visual streams of videos, enabling them to identify discrepancies within the audio-visual content. Finally, we assess the performance of these separate classifiers by applying them to the {\vadd} dataset to measure their effectiveness in detecting inconsistencies between audio and visual elements.

\subsection{Visual scene representations}
We follow a transfer learning approach for the analysis of the visual content of videos. Recognizing the quality of models pre-trained on large-scale datasets, we leverage their power to extract features that capture semantically meaningful information from input images. Initially, we experimented with several pre-trained models, including wide ResNet and DenseNet trained on the Places365 dataset, as well as an EfficientNetV2 trained on the ImageNet dataset. Aiming for a rich representation of visual information, we selected three diverse models (i.e., different network architectures pre-trained on different datasets), which are detailed in the following.

\begin{enumerate}
    \item \texttt{ViT embeddings}: We reviewed available pre-trained models and opted for the {ViT\textunderscore H\textunderscore 14} vision transformer architecture from PyTorch's Torchvision, featuring a 14x14 input patch size. We utilized the ``IMAGENET1K\textunderscore SWAG\textunderscore E2E\textunderscore V1'' weights, achieving a top-1 classification accuracy of 98.694\%. We extracted activations from the penultimate layer, resulting in a 1000-dimensional embedding vector.

    \item \texttt{CLIP embeddings}: We utilized OpenAI's CLIP architecture, which can comprehend both images and text simultaneously. Specifically, we employed the ``ViT-H/14'' architecture with ``laion2b\_s32b\_b79k'' model weights, achieving an accuracy of 78.0\% on LAION-2B. The resulting image encoding is a 1024-dimensional vector.

    \item \texttt{ResNet embeddings}: Among the various models pre-trained on the Places365 dataset available\footnote{\url{https://github.com/CSAILVision/places365}}, we selected the ResNet50 with a top-1 error rate of 44.82\%. Once again, we utilized activations from the penultimate layer, resulting in a 2048-dimensional embedding vector.
\end{enumerate}

\subsection{Audio scene representations}
Similar to the visual analysis, we employ a transfer learning approach for the analysis of the audio stream. Following the methodology outlined in \cite{Abesser_2023_Embeddings_TASLP}, we explore different deep audio embeddings (DAEs) as audio feature representations, derived from DNN models pre-trained on large-scale datasets like AudioSet \cite{Gemmeke_2017_AudioSet_ICASSP}, which covers over 2 million weakly labeled audio files across 527 sound classes.

\begin{enumerate}
    \item \texttt{OpenL3 embeddings} \cite{Cramer_2019_L3Embeddings_ICASSP}: Trained in a self-supervised fashion using an audio-visual correlation criterion during training, OpenL3 embeddings utilize separate audio and video networks, combined via fusion layers. The audio network, utilized for computing OpenL3 embeddings, consists of four convolutional layers with intermediate max-pooling operations. In this work, we use the "music" configuration, converting a Mel spectrogram with 256 Mel bands into 512-dimensional embedding vectors with a feature rate of 42 Hz.

    \item \texttt{PANN}: The Pre-trained Audio Neural Network (PANN) \cite{Kong_2021_PANNEmbeddings_TASLP} embeddings are based on a CNN architecture with 12 convolutional layers and two dense layers. As input, the ``Wavegram-Logmel-CNN'' model processes both raw audio samples and a Mel spectrogram with 64 bands. The model incorporates temporal aggregation using maximum and average pooling, resulting in 512-dimensional embedding vectors at a feature rate of 1 Hz.
    
    \item \texttt{IOV}: As a third audio representation, we obtain embeddings from a ResNet model, which has been trained for the task of indoor-outdoor-vehicle classification. We adopt the ``CNN420'' model from~\cite{Grollmisch_2021_AudioFixMatch_MDPI}, which combines a convolutional block with four residual blocks. Intermediate dropout with a rate of $0.1$ and batch normalization is used in all blocks. After the residual blocks, average pooling is applied to aggregate feature maps, with the output of the pooling layer serving as embeddings.
\end{enumerate}

\subsection{Combining modalities}
\begin{figure}[htbp]
    \centering
    \includegraphics[trim={0 17cm 18.5cm 0},clip,width=0.85\linewidth]{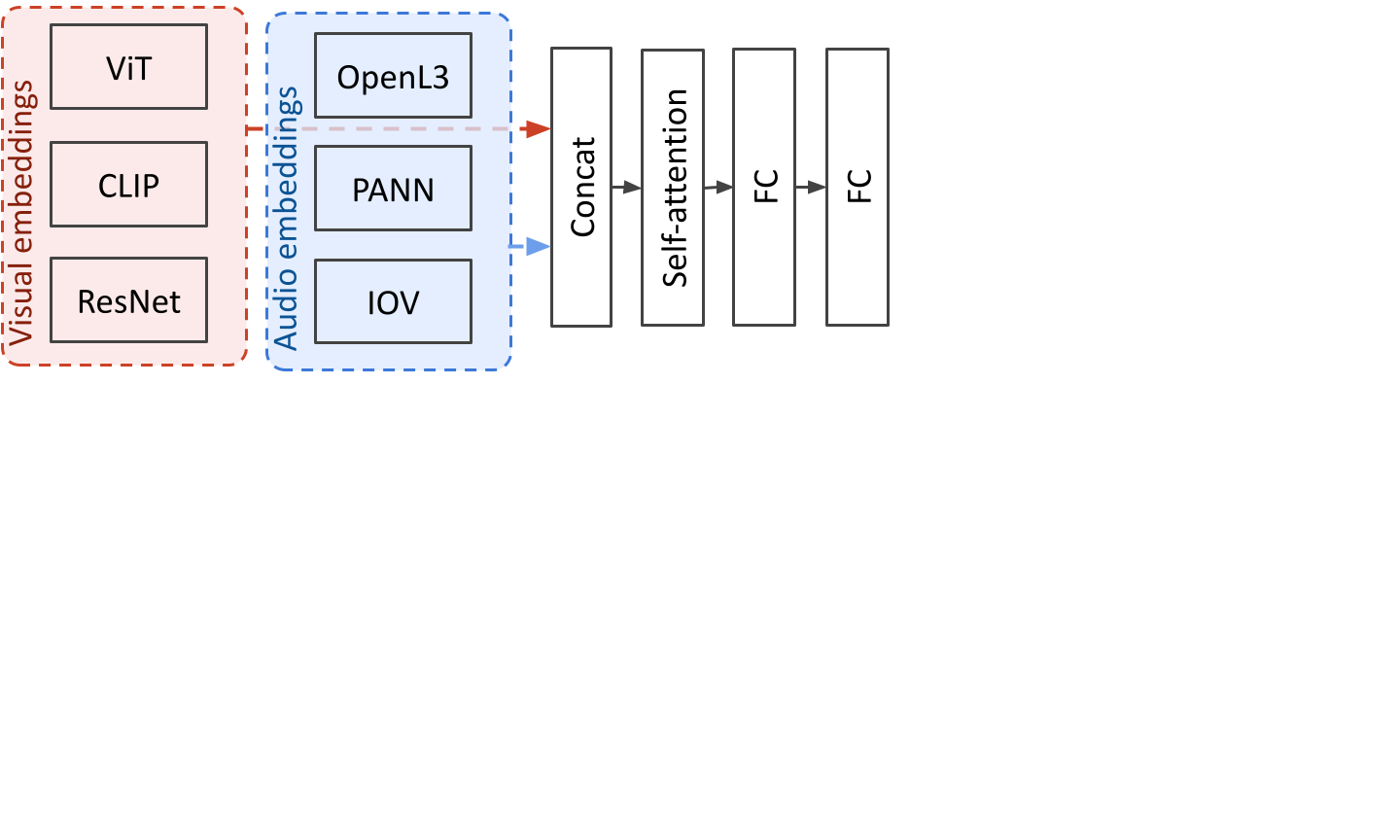}
    \caption{The architecture of the employed audio-visual scene classifier.}
    \label{fig:arch}
\end{figure}

For aggregating the three visual embeddings and three audio embeddings to classify the scene of an input video, we utilize a neural network architecture. Our approach involves concatenating these embeddings and applying a self-attention mechanism, followed by two Fully Connected (FC) layers. To prevent overfitting and enhance generalization, we include a dropout layer between the two FC layers. This straightforward architecture is illustrated in Fig.~\ref{fig:arch}.

Attention~\cite{vaswani2017attention}, in the context of deep learning, is a mechanism most commonly associated with transformer architectures; attention-based architectures have proven to be a powerful tool for both vision \cite{sun2023self} and audio \cite{hu2023joint} tasks. Unlike standard attention techniques, self-attention operates within a single sequence, capturing dependencies between elements within the same sequence. It allows a model to weigh the importance of different elements within a sequence while processing that sequence. In the context of our scene classification classifier, self-attention mechanisms are leveraged to emphasize the importance of certain parts of the different input embeddings, whether they are visual features from images or audio features from sound clips.

For the visual component, we sample the middle frame of each second in a video. Given that all videos in our training dataset are 10 seconds long, this results in 10 embeddings per video. For training purposes, we use all 10 instances of embeddings derived from the same video, all of which carry the same label. In the case of audio embeddings, we employ an averaging technique to create a single feature vector for each embedding for each second, ensuring alignment between visual and audio feature vectors. During the evaluation, we forward all 10 instances of the same video through our classifier and employ a voting scheme to infer a final classification for the video.

To enrich the training dataset and reduce training time, we employ data augmentation techniques while separating feature extraction and classification stages. We pre-compute embeddings from pre-trained models as input features for our scene classifier, significantly reducing training time. Data augmentation involves creating a duplicate training set and applying synthetic transformations only to the second half. These transformations include horizontal flips, random brightness and contrast adjustments, and rotation.

\section{Experiments}
\label{sec:exps}
Regarding the training procedure, we use a batch size of 32, as an effective balance between computational efficiency and convergence speed. We used the PyTorch machine learning library (version 1.13) and the Torchvision package utilized for accessing pre-trained models. We employ a standard stochastic gradient descent (SGD) optimizer for optimization. Learning rate scheduling is implemented, starting with a higher rate of 0.001 in the first epoch to swiftly capture prominent features. Subsequently, the rate is linearly reduced to 0.00001 by the 19th epoch to encourage meticulous fine-tuning. Training occurs over 20 epochs, as experiments showed that the loss plateaus before the 20th epoch. We selected the model snapshot from the 20th epoch and evaluated its accuracy on the test partition of the {\taud} dataset's developmental portion. We select the model snapshot from the 20th epoch and evaluate its accuracy on the test partition of the {\taud} dataset's developmental portion. All experiments are performed on a PC equipped with an Nvidia GeForce GTX 1080 Ti GPU.

\subsection{Comparison with SoA on scene classification}
We first evaluate our (joint visual-audio) scene classifier on the test portion of the development set of the {\taud} dataset to compare it with \cite{wang2021audio},  which achieved the highest performance in the DCASE 2021 audio-visual scene classification task. Our method achieves 97.24\% accuracy compared to the best score of 95.1\% reported in \cite{wang2021audio}, therefore it is clear that our method is superior, which is due to the use of modern features and, as shown in the ablation study sub-section, below, the well-chosen placing of self-attention mechanisms in our classifier architecture. 

\subsection{Visual-audio discrepancies detection}
After confirming the effectiveness of our multimodal model in scene classification using the {\taud} dataset, we proceed to apply separate visual and audio classifiers to detect discrepancies in the {\vadd} dataset. We train distinct classifiers for each modality and use them to detect the manipulated videos of the {\vadd} dataset. The evaluation of the separate classifiers on the {\vadd} dataset, both for the 3-class and 10-class variants, is included in Table~\ref{tab:exps_separate}. We observe that the audio scene and visual scene classifiers achieve near-perfect performance in the 3-class variant of {\vadd} dataset. This high accuracy suggests that our classifiers can effectively discern the intended scenes within the multimedia content. Consequently, when our model detects a discrepancy between the audio and visual scenes, it indicates a high likelihood of actual inconsistencies in the analyzed media item, minimizing the risk of false positives.

\begin{table}[htbp]
    \centering
    \caption{Evaluation of the proposed visual-audio (VASC), visual (VSC), and audio (ASC) scene classifiers on the {\taud} dataset, for the 3-class and 10-class variants.}
    \label{tab:exps_separate}
    \begin{tabular}{L{50pt}D{70pt}D{70pt}}
        \hline
        Approach & Accuracy (\%) on {\taud} using the 3-class variant  & Accuracy (\%) on {\taud} using the 10-class variant \\ \hline
        VASC  & 99.95 & 97.24 \\
        ASC   & 99.84 & 78.84 \\
        VSC   & 99.93 & 94.32 \\
        \hline
    \end{tabular}
\end{table}

\begin{figure}[htbp]
    \centering
    \includegraphics[trim={0 0 3cm 1cm},clip,width=\linewidth]{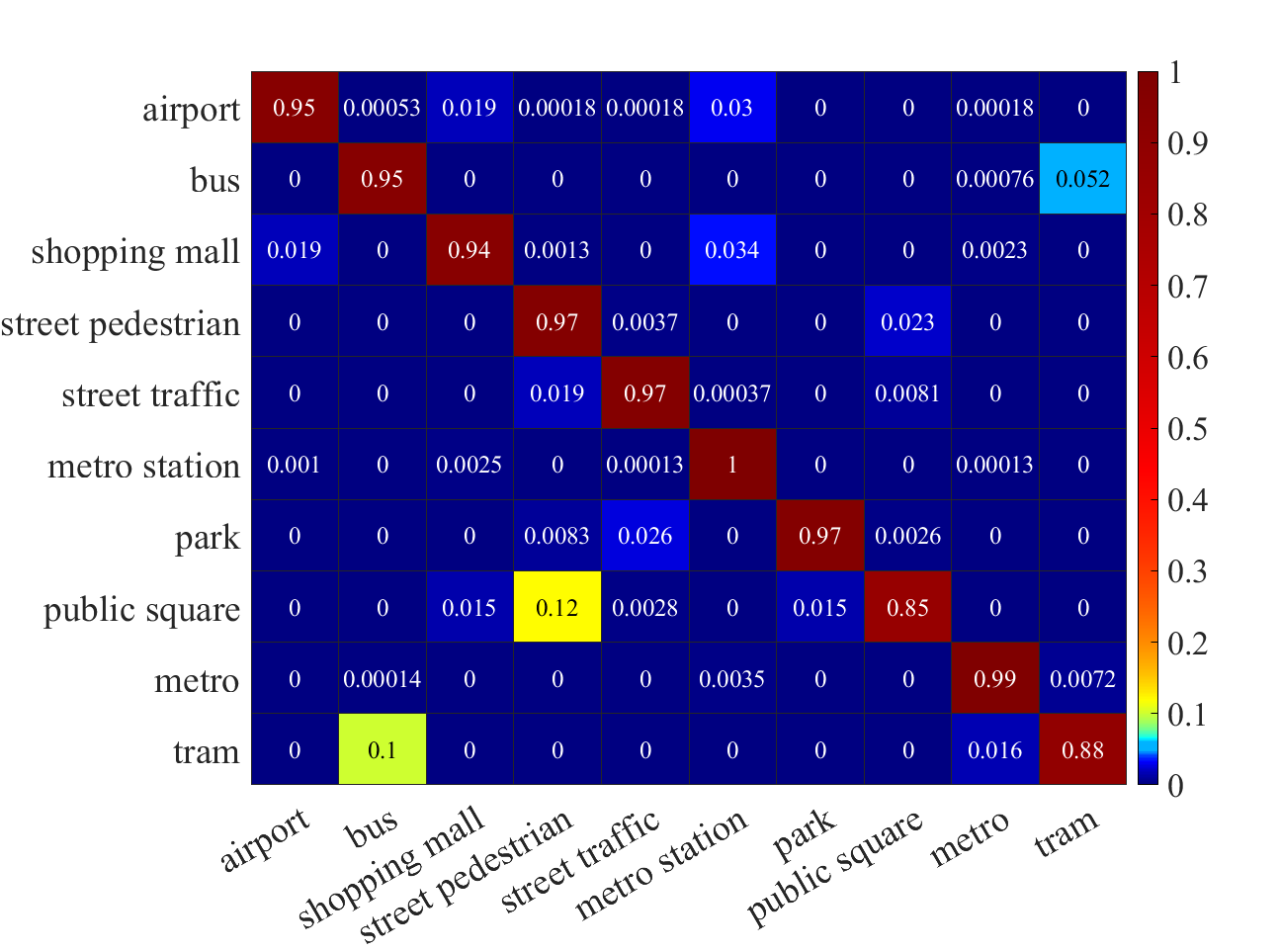}
    \caption{Confusion Matrix for our visual-audio scene classifier on the 10-class variant of the {\vadd} dataset.}
    \label{fig:cm}
\end{figure}

We evaluate the effectiveness of the proposed baseline method for detecting visual-audio discrepancies on the {\vadd} dataset, with the findings being reported in Table~\ref{tab:exps_vadd}. Our assessment reveals notable differences between the 3-class and 10-class variants. The 3-class variant demonstrates higher accuracy, achieving an F1-score of 95.54\%. This outcome is expected due to the reduced class complexity. However, it's crucial to include the 10-class variant for a more realistic and challenging evaluation, reflecting the complexities of real-world multimedia content. When applied to the 10-class variant, the baseline method achieves a lower F1-score of 79.16\%. Figure~\ref{fig:cm} displays the confusion matrix for our visual-audio scene classifier on the 10-class variant of the {\vadd} dataset, where the x-axis represents the predicted labels and the y-axis represents the true labels. A brief analysis of the results using confusion matrices showed that our classifier exhibited confusion between ``tram'' and ``bus'', ``public square'' and ``street pedestrian'', as well as ``airports" and ``metro station'' class tuples, contributing to the lower performance of our method in the 10-class variant of the problem. Researchers can leverage both variants to evaluate their detection methods across different complexity levels.

\begin{table}[htbp]
    \centering
    \caption{Comparison of the proposed baseline method on the {\vadd} dataset.}
    \label{tab:exps_vadd}
    \begin{tabular}{L{80pt}D{60pt}}
        \hline
        {\vadd} dataset variant used & F1-score (\%) of the proposed method \\ 
        \hline
        3-class {\vadd} &  95.54 \\
        10-class {\vadd} & 79.16 \\ 
        \hline
    \end{tabular}
\end{table}

\subsection{Ablation study}

\begin{figure}[htbp]
    \centering
    \includegraphics[trim={0 3.5cm 20cm 0},clip,width=0.73\linewidth]{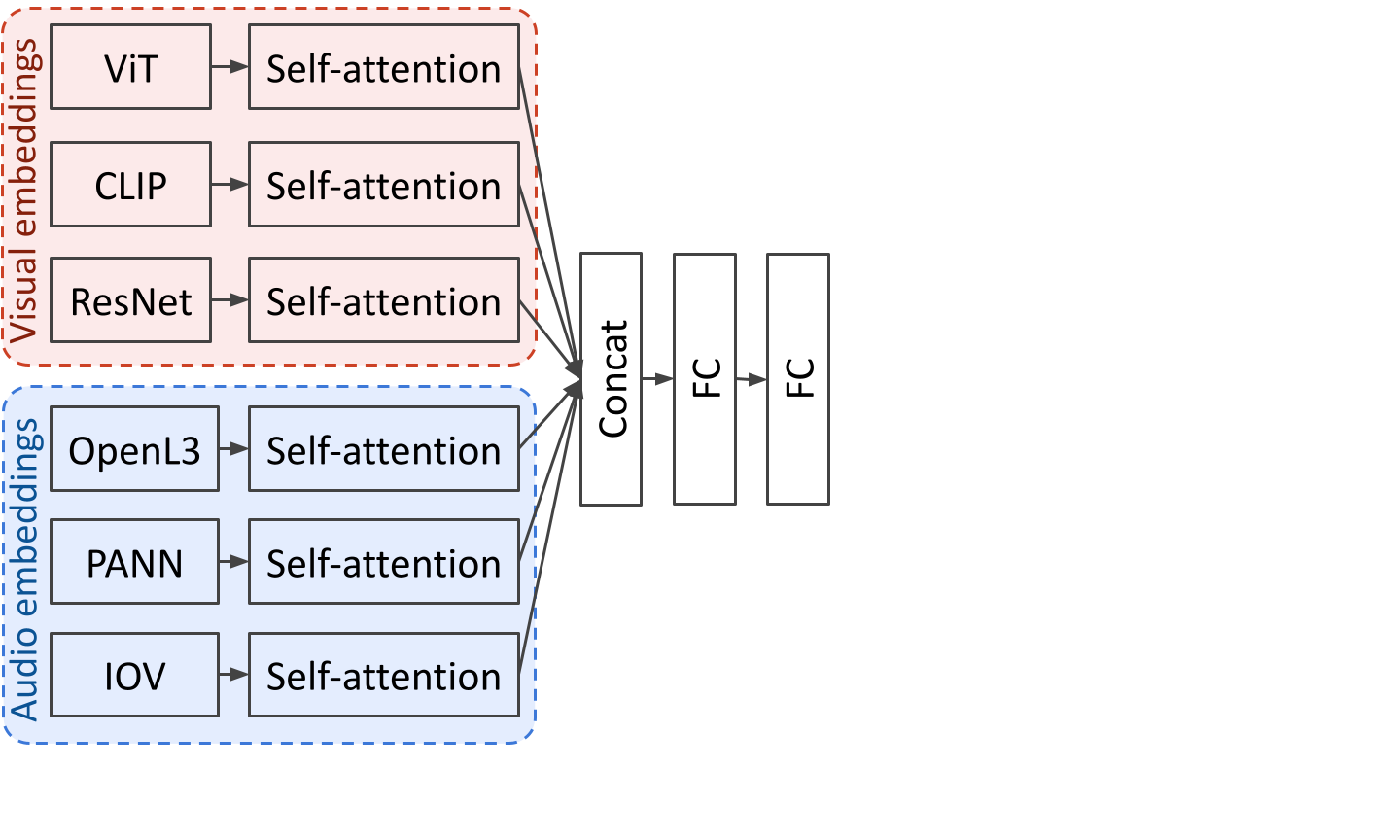}
    \caption{The architecture of the early self-attention (ES) variant.}
    \label{fig:arch_es}
\end{figure}

\begin{figure}[htbp]
    \centering
    \includegraphics[trim={0 3.5cm 25cm 0},clip,width=0.63\linewidth]{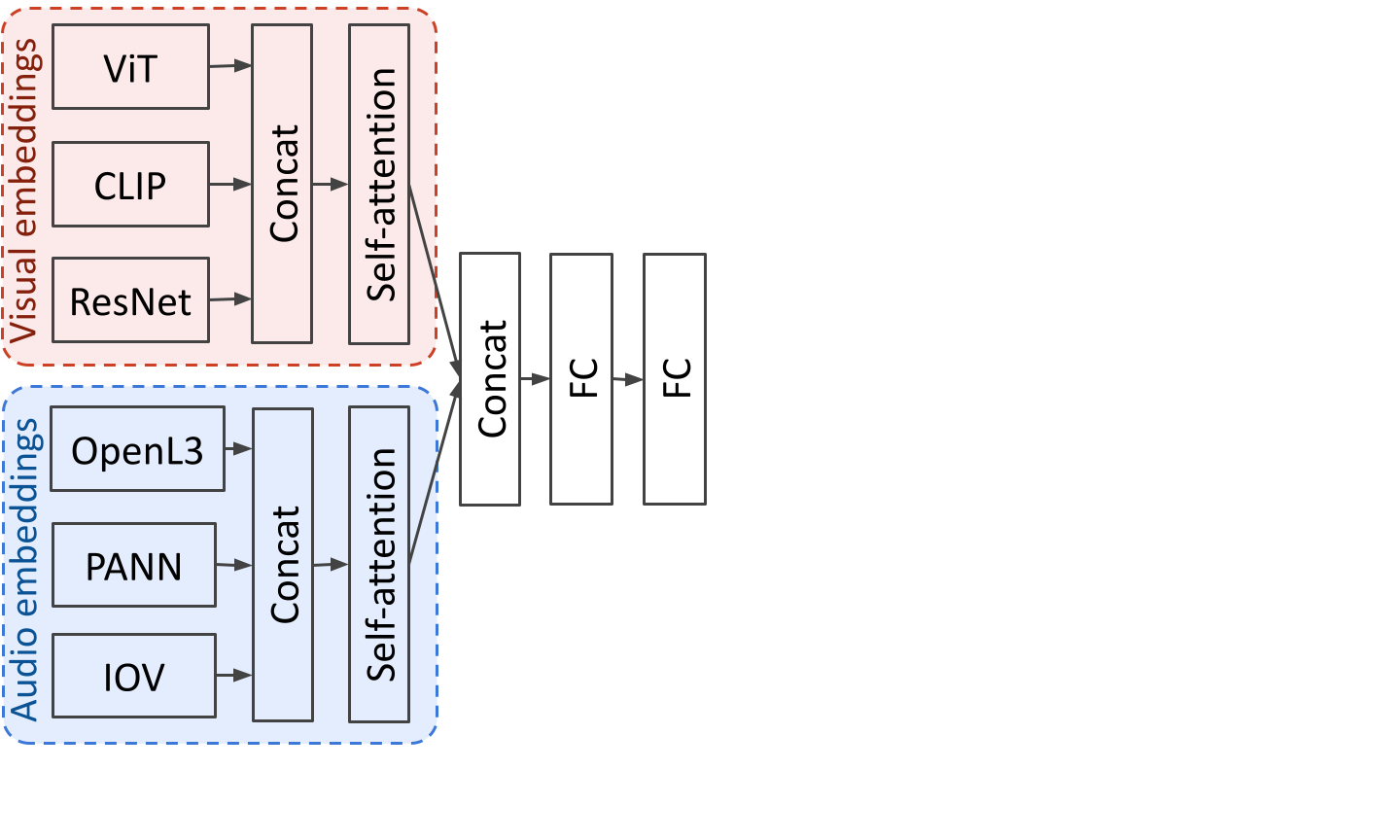}
    \caption{The architecture of the per-modality self-attention (MS) variant.}
    \label{fig:arch_ms}
\end{figure}

In this section, we conduct a brief ablation study to provide insights into the effectiveness of the core elements of our proposed model for scene classification. Our initial model combines six embeddings (three visual and three audio) to predict scene categories. We explore two primary scenarios to assess the impact of model configurations and design choices: 

\begin{enumerate}
	\item Scenario \#1 examines different variants of Self-Attention layer placement, including Late Self-Attention (LS), where self-attention is applied after concatenating all input embeddings (Fig.~\ref{fig:arch}); Early Self-Attention (ES), which applies self-attention directly to individual visual and audio embeddings before concatenation (Fig.~\ref{fig:arch_es}); Per-modality Self-Attention (MS), which applies self-attention to concatenated visual and audio embeddings (Fig.~\ref{fig:arch_ms}); Combined Self-Attention, which encompasses various combinations of ES, MS, and LS approaches; and finally, not using self-attention at all (NS).
	\item Scenario \#2 investigates Data Augmentation (DA) techniques and their impact on scene classification performance through experiments both with and without augmentation. 
    \item Scenario \#3 - Single vs. Double FC Layers: we aim to determine whether the additional FC layer contributes significantly to the model’s predictive power.
\end{enumerate}

While we present only the ablation study results using the 3-class variant of the {\vadd} dataset, it's important to note that the findings remain applicable to the 10-class variant as well.

\begin{table}[htbp]
    \centering
    \caption{Comparison of different design options for the proposed classifier}
    \label{tab:exps_ablation}
    \begin{tabular}{L{50pt}L{80pt}D{70pt}}
        \hline
        & Approach & Accuracy (\%) on {\taud} dataset \\
        \hline
        & Proposed (LS + DA + + Double FC) & \textbf{97.24} \\ \hline
       \multirow{7}{*}{\begin{tabular}[c]{@{}l@{}}Scenario \#1:\\ Self-attention\\ variants\\ (all using DA)\end{tabular}}
        & ES & 91.73 \\
        & MS & 96.98 \\ 
        & NS & 94.16 \\
        & ES + LS & 93.75 \\
        & MS + LS & 97.02 \\
        & ES + MS & 92.65 \\
        & ES + MS + LS & 94.05 \\ \hline
        Scenario \#2 & Proposed without DA & 96.98 \\ \hline
        Scenario \#3 & Single FC layer & 97.18 \\
        \hline
    \end{tabular}
\end{table}

The outcome of the ablation study is reported in Table~\ref{tab:exps_ablation}. Regarding the position of a self-attention mechanism in the model, its placement can significantly influence the model's ability to capture relevant information and dependencies within the input data. Placing self-attention early in the model allows the network to attend to fine-grained features at lower levels of abstraction, potentially enhancing feature learning. On the other hand, positioning self-attention later in the model enables the network to integrate contextual information and global dependencies, facilitating better aggregation of information from multiple input sources. In our case, positioning the self-attention late in the network works best for the embedding aggregation scheme (LS $>$ [ES, MS]). Furthermore, overloading the network with more than one self-attention layer seems to ``short-circuit'' the model, leading to significantly reduced accuracy (LS, ES, MS $>$ [ES+LS, MS+LS, ES+MS, ES+MS+LS]). Data augmentation schemes benefit training by increasing the diversity and richness of the training data, thereby improving the model's ability to generalize to unseen examples. Employing a data augmentation scheme increases accuracy on the specific task. Finally, the performance improvement observed when using two fully connected (FC) layers instead of one can be attributed to the increased capacity and expressiveness of the deeper network architecture. Our experiments across all scenarios consistently demonstrate that the chosen model architecture, data augmentation strategies, and the number of FC layers used, are well-suited for the task at hand.

\section{Conclusions}
\label{sec:conclusions}
In this paper, we introduce a baseline method that utilizes multi-modal scene classification techniques for content verification, specifically focusing on identifying inconsistencies between audio and visual elements in videos. Our approach involves developing and evaluating a robust joint audio-visual classifier using the existing {\taud} dataset, demonstrating its effectiveness in scene classification. We extend this classifier to the context of content verification, providing a valuable tool for assessing media integrity. Additionally, we introduce a benchmark dataset to facilitate further research in this area. The evaluation of our separate classifiers on the newly introduced dataset reveals promising results in the 3-class variant, while also highlighting current limitations in the 10-class variant of the proposed experimental protocol. In conclusion, our research aims to advance the detection of audio-visual discrepancies, offering valuable resources and insights for future studies. We intend to look into different feature fusion strategies and contrastive learning approaches as well as incorporating temporal information so that we capture discrepancies over time, as our next steps. 

\begin{acks}
This work was supported by the EU Horizon Europe programme under grant agreement 101070093 vera.ai.
\end{acks}

%\bibliographystyle{ACM-Reference-Format}
%\bibliography{refs}


\end{document}